\documentclass[10pt,twocolumn,letterpaper]{article}

\usepackage{iccv}
\usepackage{times}
\usepackage{epsfig}
\usepackage{graphicx}
\usepackage{amsmath}
\usepackage{amssymb}
\usepackage{booktabs}
\usepackage{subfigure}
\usepackage{enumitem}
\usepackage[toc,page]{appendix}

\usepackage{booktabs}
\usepackage{pifont}

\newlength\savewidth\newcommand\shline{\noalign{\global\savewidth\arrayrulewidth
  \global\arrayrulewidth 1pt}\hline\noalign{\global\arrayrulewidth\savewidth}}

\usepackage[table,x11names, dvipsnames]{xcolor}

\usepackage[pagebackref=true,breaklinks=true,letterpaper=true,colorlinks,bookmarks=false]{hyperref}

\iccvfinalcopy 

\def\paperTitle{DFA3D: 3D Deformable Attention For 2D-to-3D Feature Lifting}
\def\methodname{{DFA3D}}

\def\authorBlock{
    \textbf{Hongyang Li}$^{1,3}$\thanks{This work was done during the internship at IDEA. } \thanks{Equal contribution. }  \qquad
    \textbf{Hao Zhang}$^{2,3}$\footnotemark[1] \footnotemark[2]  \qquad
    \textbf{Zhaoyang Zeng}$^{3}$ \qquad
    \textbf{Shilong Liu}$^{4,3}$ \\
    \textbf{Feng Li}$^{2,3}$\qquad
    \textbf{Tianhe Ren}$^{3}$\qquad
    \textbf{Lei Zhang}$^{1,3}$\thanks{Corresponding author.} \\
    $^1$South China University of Technology. \\
    $^2$The Hong Kong University of Science and Technology. \\
    $^3$International Digital Economy Academy (IDEA). \\
    $^4$Dept. of CST., BNRist Center, Institute for AI, Tsinghua University. \\
}

\ificcvfinal\fi

\begin{document}

\title{\paperTitle}

\author{\authorBlock}

\maketitle
\ificcvfinal\thispagestyle{empty}\fi


\begin{abstract}
In this paper, we propose a new operator, called 3D DeFormable Attention (DFA3D), for 2D-to-3D feature lifting, which transforms multi-view 2D image features into a unified 3D space for 3D object detection. 
Existing feature lifting approaches, such as Lift-Splat-based and 2D attention-based, either use estimated depth to get pseudo LiDAR features and then splat them to a 3D space, which is a one-pass operation without feature refinement, or ignore depth and lift features by 2D attention mechanisms, which achieve finer semantics while suffering from a depth ambiguity problem. 
In contrast, our DFA3D-based method first leverages the estimated depth to expand each view's 2D feature map to 3D and then utilizes DFA3D to aggregate features from the expanded 3D feature maps. With the help of DFA3D, the depth ambiguity problem can be effectively alleviated from the root, and the lifted features can be progressively refined layer by layer, thanks to the Transformer-like architecture. In addition, we propose a mathematically equivalent implementation of DFA3D which can significantly improve its memory efficiency and computational speed. We integrate DFA3D into several methods that use 2D attention-based feature lifting with only a few modifications in code and evaluate on the nuScenes dataset. The experiment results show a consistent improvement of +1.41\% mAP on average, and up to +15.1\% mAP improvement when high-quality depth information is available, demonstrating the superiority, applicability, and huge potential of DFA3D. The code is available at \href{https://github.com/IDEA-Research/3D-deformable-attention.git}{https://github.com/IDEA-Research/3D-deformable-attention.git}.
\end{abstract}

\section{Introduction}
3D object detection is a fundamental task in many real-world applications such as robotics and autonomous driving. Although LiDAR-based methods~\cite{yin2021center, zhou2018voxelnet, pan20213d, wang2021object} have achieved impressive results with the help of accurate 3D perception from LiDAR, multi-view camera-based methods have recently received extensive attention because of their low cost for deployment and distinctive capability of long-range and color perception. Classical multi-view camera-based 3D object detection approaches mostly follow monocular frameworks which first perform 2D/3D object detection in each individual view and then conduct cross-view post-processing to obtain the final result. While remarkable progress has been made~\cite{wang2022probabilistic,wang2021fcos3d,park2021pseudo,reiher2020sim2real,bruls2019right}, such a framework cannot fully utilize cross-view information and usually leads to low performance.

To eliminate the ineffective cross-view post-processing, several end-to-end approaches~\cite{philion2020lift,li2022bevdepth,huang2021bevdet, li2022bevformer, wang2022detr3d,liu2022petr,liu2022petrv2} have been developed. These approaches normally contain three important modules: a backbone for 2D image feature extraction, a feature lifting module for transforming multi-view 2D image features into a unified 3D space (e.g. BEV space in an ego coordinate system) to obtain lifted features, and a detection head for performing object detection by taking as input the lifted features. Among these modules, the feature lifting module serves as an important component to bridge the 2D backbone and the 3D detection head, whose quality will greatly affect the final detection performance.

\begin{figure*}[t]
    \centering
        \includegraphics[width=0.95\linewidth]{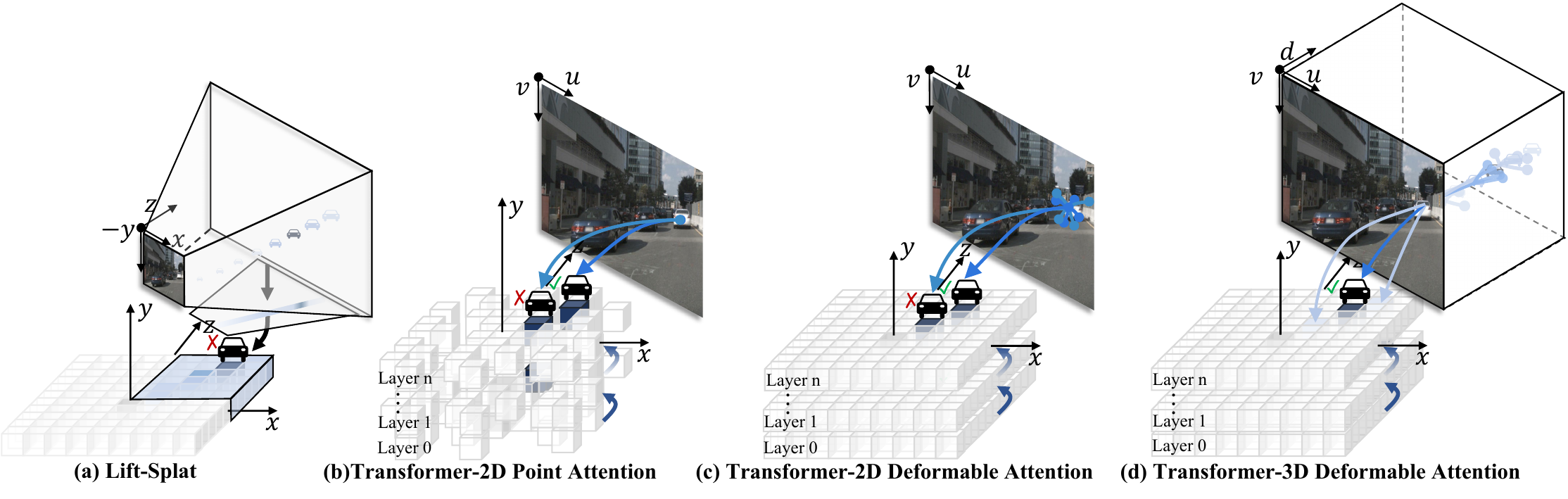}
    \caption{Comparisons of feature lifting methods. (a) Lift-Splat-based methods cannot handle depth errors after feature lifting, which can result in wrong predictions. Point attention-based (b) and 2D deformable attention-based (c) can refine the lifted features layer-by-layer. However, they suffer from depth ambiguity when multiple 3D objects are projected to the same 2D point, which may result in duplicate predictions along a ray connecting the ego car and a target object.
    (d) Our 3D deformable attention can effectively alleviate the depth ambiguity problem as we sample features in 3D spaces. 
    Moreover, the multi-layer design of Transformers helps refine features and sampling points layer by layer.
    } 
    \label{fig.intro}
    \vspace{-0.6cm}
\end{figure*}

To perform feature lifting, recent methods usually predefine a set of 3D anchors in the ego coordinate system sparsely or uniformly, with randomly initialized content features. After that, they lift 2D image features into the 3D anchors to obtain the lifted features.
Some methods~\cite{li2022bevdepth,huang2021bevdet,huang2022bevdet4d,wang2022sts,li2022bevstereo} utilize a straightforward lift and splat mechanism~\cite{philion2020lift} by first lifting 2D image features into pseudo LiDAR features in an ego coordinate system using estimated depth and then assigning the pseudo LiDAR features to their closest 3D anchors to obtain the final lifted features, as shown in Fig.~\ref{fig.intro}(a). Although such methods have achieved remarkable performance, due to their excessive resource consumption when lifting 2D image features, they normally cannot utilize multi-scale feature maps, which are crucial for detecting small (faraway) objects. 

Instead of lift and splat, some other works utilize 2D attention to perform feature lifting. Such methods treat predefined 3D anchors as 3D queries and propose several ways to aggregate 2D image features (keys) to 3D anchors. Typical 2D attention mechanisms that have been explored include dense attention ~\cite{liu2022petr}, point attention (a degenerated deformable attention) ~\cite{detr,chen2022graph}, and deformable attention ~\cite{li2022bevformer,yang2022bevformer,chu2023oa}. 2D attention-based methods can naturally work with Transformer-like architectures to progressively refine the lifted features layer by layer. Moreover, deformable attention makes it possible to use multi-scale 2D features, thanks to its introduced sparse attention computation. 
However, the major weakness of 2D attention-based approaches is that they suffer from a crucial depth ambiguity problem, as shown in Fig.~\ref{fig.intro}(b,c). That is, due to the ignorance of depth information, when projecting 3D queries to a 2D view, many 3D queries will end up with the same 2D position with similar sampling points in the 2D view. This will result in highly entangled aggregated features and lead to wrong predictions along a ray connecting the ego car and a target object (as visualized in Fig.~\ref{fig.visualization}).
Although some concurrent works~\cite{lin2022sparse4d, dabev} have tried to alleviate this issue 
, they can not fix the problem from the root. 
The root cause lies in the lack of depth information when applying 2D attention to sample features in a 2D pixel coordinate system\footnote{The pixel coordinate system that only considers the $u,v$ axes.}.

To address the above problems, in this paper, we propose a basic operator called 3D deformable attention (DFA3D), built upon which we develop a novel feature lifting approach, as shown in Fig. \ref{fig.intro}(d). We follow \cite{philion2020lift} to leverage a depth estimation module to estimate a depth distribution for each 2D image feature. We expand the dimension of each single-view 2D image feature map by computing the outer product of them and their estimated depth distributions to obtain the expanded 3D feature maps in a 3D pixel coordinate system\footnote{The pixel coordinate system that considers all of the $u,v,d$ axes.}. Each 3D query in the BEV space is projected to the 3D pixel space with a set of predicted 3D sampling offsets to specify its 3D receptive field and pool features from the expanded 3D feature maps. 
In this way, a 3D query close to the ego car only pools image features with smaller depth values, whereas a 3D query far away from the ego car mainly pools image features with larger depth values, and thus the depth ambiguity problem is effectively alleviated.

In our implementation, as each of the expanded 3D feature maps is an outer product between a 2D image feature map and its estimated depth distributions, we do not need to maintain a 3D tensor in memory to avoid excessive memory consumption. Instead, we seek assistance from math to simplify DFA3D into a mathematically equivalent depth-weighted 2D deformable attention and implement it through CUDA, making it both memory-efficient and fast. The Pytorch interface of DFA3D is very similar to 2D deformable attention and only requires a few modifications to replace 2D deformable attention, making our feature lifting approach easily portable.

In summary, our main contributions are:
\begin{enumerate}[itemsep=0pt, topsep=0pt, parsep=0pt]

    \item We propose a basic operator called 3D deformable attention (DFA3D) for feature lifting. Leveraging the property of outer product between 2D features and their estimated depth, we develop a memory-efficient and fast implementation. 
    
    \item Based on DFA3D, we develop a novel feature lifting approach, which not only alleviates the depth ambiguity problem from the root, but also benefits from multi-layer feature refinement of a Transformer-like architecture.
    Thanks to the simple interface of DFA3D, 
    our DFA3D-based feature lifting can be implemented in any method that utilizes 2D deformable attention (also its degeneration)-based feature lifting with only a few modifications in code.
    
    \item The consistent performance improvement (+1.41 mAP on average) in comparative experiments on the nuScenes~\cite{caesar2020nuscenes} dataset demonstrate the superiority and generalization ability of DFA3D-based feature lifting.
\end{enumerate}
  
\section{Related Work}

\noindent
\textbf{Multi-view 3D Detectors with Lift-Splat-based Feature Lifting: }
To perform 3D tasks in the multi-view context in an end-to-end pipeline, Lift-Splat~\cite{philion2020lift} proposes to first lift multi-view image features by performing outer product between 2D image feature maps and their estimated depth, and then transform them to a unified 3D space through camera parameters to obtain the lifted pseudo LiDAR features. Following the LiDAR-based methods, the pseudo LiDAR features are then splatted to predefined 3D anchors through a voxel-pooling operation to obtain the lifted features. The lifted features are then utilized to facilitate the following downstream tasks. For example, BEVDet~\cite{huang2021bevdet} proposes to adopt such a pipeline for 3D detection and verifies its feasibility. Although being effective, such a method suffers from the problem of huge memory consumption, which prevents them from using mutli-scale feature maps. Moreover, due to the fixed assignment rule between the pseudo LiDAR features and the 3D anchors, each 3D anchor can only interact with image features once without any further adjustment later. This problem makes Lift-Splat-based methods reliant on the quality of depth. The recently proposed BEVDepth~\cite{li2022bevdepth} makes a full analysis and validates that the learned depth quality has a great impact on performance. As the implicitly learned depth in previous works does not satisfy the accuracy requirement, BEVDepth proposes to resort to monocular depth estimation for help. By explicitly supervising monocular depth estimation with ground truth depth obtained by projecting LiDAR points onto multi-view images, BEVDepth achieves a remarkable performance. Inspired by BEVDepth, instead of focusing on the above-discussed problems of refinement or huge memory consumption, many works~\cite{li2022bevstereo,wang2022sts} try to improve their performance through improving depth quality by resorting to multi-view stereo or temporal information. However, due to small overlap regions between multi-view images and also the movements of surrounding objects, they still rely on the monocular depth estimation, which is an ill-posed problem and can not be accurately addressed.

\noindent
\textbf{Multi-view 3D Detectors with 2D Attention-based Feature Lifting: }
Motivated by the progress in 2D detection~\cite{ren2023detrex, ren2023strong, liu2023detection, liu2023grounding, zhang2022dino, li2022dn, liu2022dab}, many works propose to introduce attention into camera-based 3D detection. They treat multi-view image features as keys and values and the predefined 3D anchors as 3D queries in the unified 3D space. The 3D queries are projected onto multi-view 2D image feature maps according to the extrinsic and intrinsic camera parameters and lift features from the image feature maps through cross attention like the decoder part of 2D detection methods~\cite{detr,deformable_detr}. 
PETR~\cite{liu2022petr} and PETRv2~\cite{liu2022petrv2} are based on the classical (dense) attention mechanism, where the interactions between each 3D query and all image features (keys) make it inefficient. 
DETR3D-like methods~\cite{wang2022detr3d, chen2022graph} proposes to utilize the point attention which lets a query interact with only one key obtained by bilinear interpolation according to its projected location (also called point deformable attention~\cite{li2022bevformer}). Although being efficient, the point attention has a small receptive field in multi-view 2D image feature maps and normally results in a relatively weak performance. 
By contrast, BEVFormer~\cite{li2022bevformer, yang2022bevformer} proposes to apply 2D deformable attention to let each 3D query interact with a local region around the location where the 3D query is projected to and obtains a remarkable result. OA-BEV~\cite{chu2023oa} tries to introduce 2D detection into the pipeline to guide the network to further focus on the target objects. 
Benefiting from the Transformer-like architecture, the lifted features obtained by 2D attention can be progressively refined layer-by-layer.
However, the methods with 2D attention-based feature lifting suffer from the problem of depth ambiguity: when projecting 3D queries onto a camera view, those with the same projected coordinates $(u, v)$ but different depth values $d$ end up with the same reference point and similar sampling points in the view. Consequently, these 3D queries will aggregate features that are strongly entangled and thus result in duplicate predictions along a ray connecting the ego car and a target object in BEV. 
Some concurrent works~\cite{dabev,lin2022sparse4d} have noticed this problem, but none of them tackle the problem from the root.

Instead of conducting feature lifting using existing operators, such as voxel-pooling and 2D deformable attention, we develop a basic operator DFA3D to perform feature lifting. As summarized in Table~\ref{tab:compare_rel}, our proposed DFA3D-based feature lifting can not only benefit from multi-layer refinement but also address the depth ambiguity problem. Moreover, DFA3D-based feature lifting can take multi-scale feature maps into consideration efficiently.

\begin{table}[b]
\vspace{-7mm}
\centering
\renewcommand\arraystretch{1.2}
\caption{Comparison with other feature lifting methods. 
$^\dag$ indicates the direct utilization of multi-scale feature maps without sampling.}
\vspace{1mm}
\resizebox{1\linewidth}{!}{
\setlength{\tabcolsep}{1.5pt}
\begin{tabular}{l|c|c|c} 
\shline
Methods           & depth differentiation & multi-layer refinement & multi-scale$^\dag$ \\
\shline
Lift-Splat-based         & \checkmark           &                        &   \\
2D Attention-based       &                      & \checkmark             & \checkmark \\
DFA3D-based (ours)\               & \checkmark           & \checkmark             & \checkmark \\

\shline
\end{tabular}
}
\vspace{-2mm}
\label{tab:compare_rel}
\end{table}

\begin{figure*}[t]
    \vspace{-0.1mm}
    \centering
        \includegraphics[width=1\linewidth]{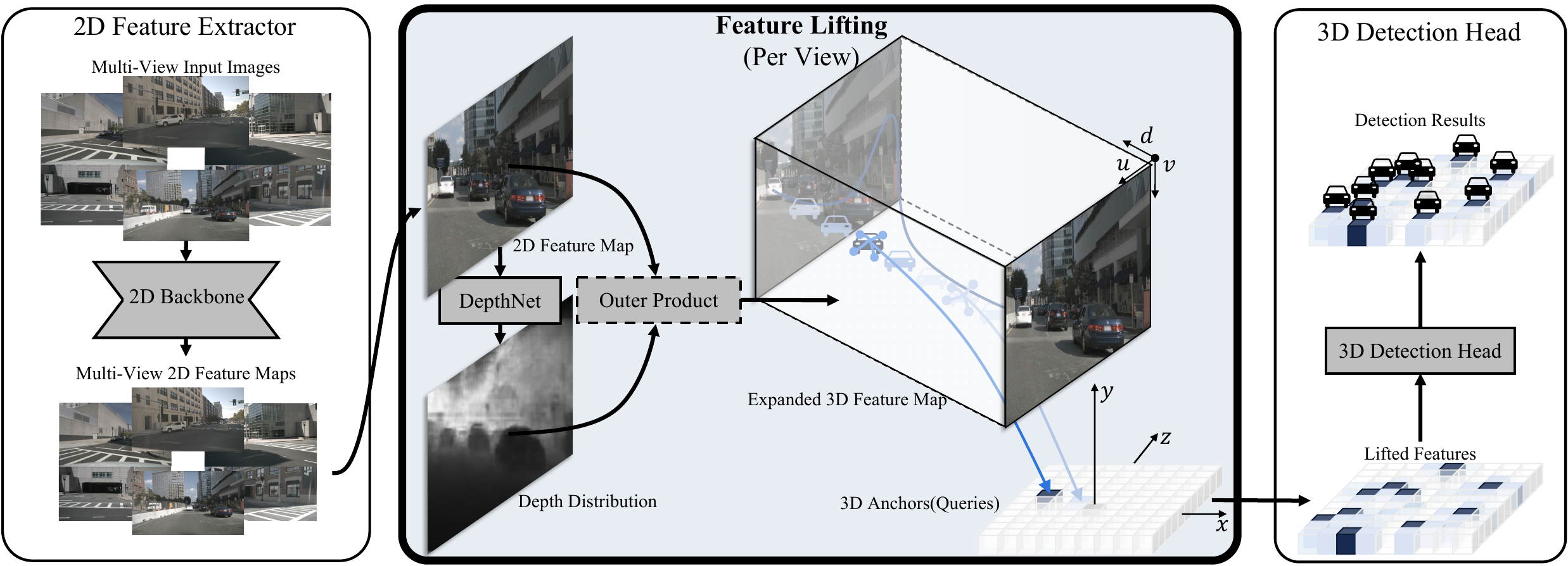}
    \caption{The overview of an end-to-end multi-view 3D detection
pipeline with {\methodname}. The 2D feature extractor takes multi-view images as inputs and extracts multi-scale 2D image feature maps for each view independently. After that, the 2D image feature maps are fed into {\methodname}-based feature lifting process and lifted to predefined 3D anchors, which are considered as 3D queries during the process. Finally, a 3D detection head takes the lifted features as input and predicts 3D bounding boxes as the final detection results.
The ``Outer Product" enclosed by a dashed border indicates that we do not actually conduct it. For better visualization, we assume the feature maps have one scale. We present 2D image feature maps by RGB images and depth distributions by a gray-scale depth map. Best viewed in color.}
    \label{fig.pipeline}
\end{figure*}

\section{Approach}

The overview of an end-to-end multi-view 3D detection approach with DFA3D-based feature lifting is shown in Fig.~\ref{fig.pipeline}. In this section, we introduce how to construct the expanded 3D feature maps for DFA3D in Sec.~\ref{subsec.feature_expanding} and how DFA3D works to realize feature lifting in Sec.~\ref{subsec.3DDeformable}. In addition, we explain our efficient implementation of DFA3D in Sec.~\ref{sec.3D_attention_impl}. Finally, we provide a detailed analysis and comparison between our DFA3D-based feature lifting and the current mainstream feature lifting methods in Sec.~\ref{subsec.analysis}.

\subsection{Feature Expanding}
\label{subsec.feature_expanding}

In the context of multi-view 3D detection, the depth information of input images is unavailable.  We follow~\cite{li2022bevdepth, wang2022sts, li2022bevstereo, huang2021bevdet} to introduce a monocular depth estimation to supplement depth information for 2D image features. Specifically, we 
follow~\cite{li2022bevdepth} to adopt a DepthNet~\cite{fu2018deep} module, which takes as input the 2D image feature maps from the backbone and generates a discrete depth distribution for each 2D image feature. The DepthNet module can be trained by using LiDAR information~\cite{li2022bevdepth} or ground truth 3D bounding box information~\cite{chu2023oa} as supervision.

We denote the multi-view 2D image feature maps as $\boldsymbol{X}\in \mathbb{R}^{V\times H\times W\times C}$, where $V$, $H$, $W$, and $C$ indicates the number of views, spatial height, spatial width, and the number of channels, respectively. We feed the multi-view 2D image feature maps into DepthNet to obtain the discrete depth distributions. The distributions have the same spatial shape as $\boldsymbol{X}$ and can be denoted as $\boldsymbol{D}\in \mathbb{R}^{V\times H\times W\times D}$, where $D$ indicates the number of pre-defined discretized depth bins.
Here, for simplicity, we assume that $\boldsymbol{X}$ only has one feature scale. In the context of multi-scale, in order to maintain the consistency of depths across multiple scales, we select the feature maps of one scale to participate in the generation of depth distributions. The generated depth distributions are then interpolated to obtain the depth distributions for multi-scale feature maps.

After obtaining the discrete depth distributions $\boldsymbol{D}$, we expand the dimension of $\boldsymbol{X}$ into 3D by conducting the outer product between $\boldsymbol{D}$ and $\boldsymbol{X}$, which is formulated as $\boldsymbol{F} = \boldsymbol{D} \otimes_{-1} \boldsymbol{X}$,
where $\otimes_{-1}$ indicates the outer product conducted at the last dimension, and $\boldsymbol{F}\in \mathbb{R}^{V\times H\times W\times D\times C}$ denotes the multi-view expanded 3D feature maps. For more details about the expanded 3D feature maps, please refer to the further visualization in appendix.

Directly expanding 2D image feature maps through outer products will lead to high memory consumption, especially when taking multi-scale into consideration. To address this issue, we develop a memory-efficient algorithm to solve this issue, which will be explained in Sec.~\ref{sec.3D_attention_impl}.

\subsection{3D Deformable Attention and Feature Lifting}
\label{subsec.3DDeformable}
After obtaining the multi-view expanded 3D feature maps, DFA3D is utilized as the backend to transform them into a unified 3D space to obtain the lifted features. Specifically, in the context of feature lifting, DFA3D treats the predefined 3D anchors as 3D queries, the expanded 3D feature maps as 3D keys and values, and performs deformable attention in the 3D pixel coordinate system.

To be more specific, for a 3D query located at $(x,y,z)$ in the ego coordinate system, its view reference point $\boldsymbol{R}_n=(u_n,v_n,d_n)$ in the 3D pixel coordinate system for the $n$-th view is calculated by\footnote{Since there is only a rigid transformation between the camera coordinate system of $n$-th view and the ego coordinate system, we assume the extrinsic matrix as identity matrix and drop it for notation simplicity.}
\begin{equation}
\begin{aligned}
d_n
\begin{bmatrix}
u_n\\
v_n\\
1\\
\end{bmatrix}
&=
\begin{bmatrix}
f_n^x & 0 & u_n^0\\
0 & f_n^y & v_n^0\\
0 & 0     & 1
\end{bmatrix} 
\begin{bmatrix}
x\\
y\\
z\\
\end{bmatrix},
\end{aligned}
\label{equ.cam2img}
\end{equation}
where $f_n^x$, $f_n^y$, $u_n^0$, $v_n^0$ are the corresponding camera intrinsic parameters. By solving Eq.~\ref{equ.cam2img}, we can obtain

\begin{equation}
\begin{aligned}
u_n = f_n^x\frac{x}{d_n} + u_n^0, v_n = f_n^y\frac{y}{d_n} + v_n^0, d_n = z
\end{aligned}.
\label{equ.sameuv}
\end{equation}

We denote the content feature of a 3D query as $\boldsymbol{q} \in \mathbb{R}^{C_q}$, where $C_q$ indicates the dimension of $\boldsymbol{q}$. The 3D deformable attention mechanism in the $n$-th view can be formulated by\footnote{Here we set the number of attention heads as 1 and the number of feature scales as 1, and drop their indices for notation simplicity.}
\begin{equation}
    \begin{aligned}
        \Delta \boldsymbol{S} &= \boldsymbol{W}^{S}\boldsymbol{q}, \Delta \boldsymbol{S}_k = \Delta \boldsymbol{S}_{[3k:3(k+1)]}, \boldsymbol{A} = \boldsymbol{W}^{A}\boldsymbol{q},\\
        \boldsymbol{q}_n &= \sum_{k=1}^{K}{\boldsymbol{A}_k\operatorname{Trili}\left(\boldsymbol{F}_n,
\boldsymbol{R}_n
+\Delta \boldsymbol{S}_k\right)},
    \end{aligned}
\label{equ.3ddef}
\end{equation}
where $\boldsymbol{q}_n$ is the querying result in the $n$-th view, $K$ denotes the number of sampling points, $\boldsymbol{W}^S \in \mathbb{R}^{3K\times C_q}$ and $\boldsymbol{W}^A \in \mathbb{R}^{K\times C_q}$ are learnable parameters, $\Delta \boldsymbol{S}$ and $\boldsymbol{A}$ represent the view sampling offsets in the 3D pixel coordinate system and their corresponding attention weights respectively. $\operatorname{Trili}(\cdot)$ indicates the trilinear interpolation used to sample features in the expanded 3D feature maps. The detailed implementation of $\operatorname{Trili}(\cdot)$ and our simplification will be explained in Sec.~\ref{sec.3D_attention_impl}.

After obtaining the querying results through DFA3D in multiple views, the final lifted feature $\boldsymbol{q}'$ for the 3D query is obtained by  query result aggregation, 
\begin{equation}
\begin{aligned}
    \boldsymbol{q}' = \sum_{n} \boldsymbol{V}_n \boldsymbol{q}_n
\end{aligned},
\label{equ.3ddef_mv}
\end{equation}
where $\boldsymbol{V}_n$ indicates the visibility of the 3D query in the $n$-th view. The lifted feature is used to update the content feature of the 3D query for feature refinement.

\subsection{Efficient 3D Deformable Attention}\label{sec.3D_attention_impl}

Although the 3D deformable attention has been formulated in Eq. \ref{equ.3ddef}, maintaining the multi-view expanded 3D feature maps $\boldsymbol{F}\in \mathbb{R}^{V\times H\times W\times D\times C}$ will lead to tremendous memory consumption, especially when taking multi-scale features into consideration. We prove that in the context of feature expanding through the outer product between 2D image feature maps and their discrete depth distributions, the trilinear interpolation can be transformed into a depth-weighted bilinear interpolation. Hence the 3D deformable attention can also be transformed into a depth-weighted 2D deformable attention accordingly.

\begin{figure}[t]
    \centering
	\includegraphics[width=1\linewidth]{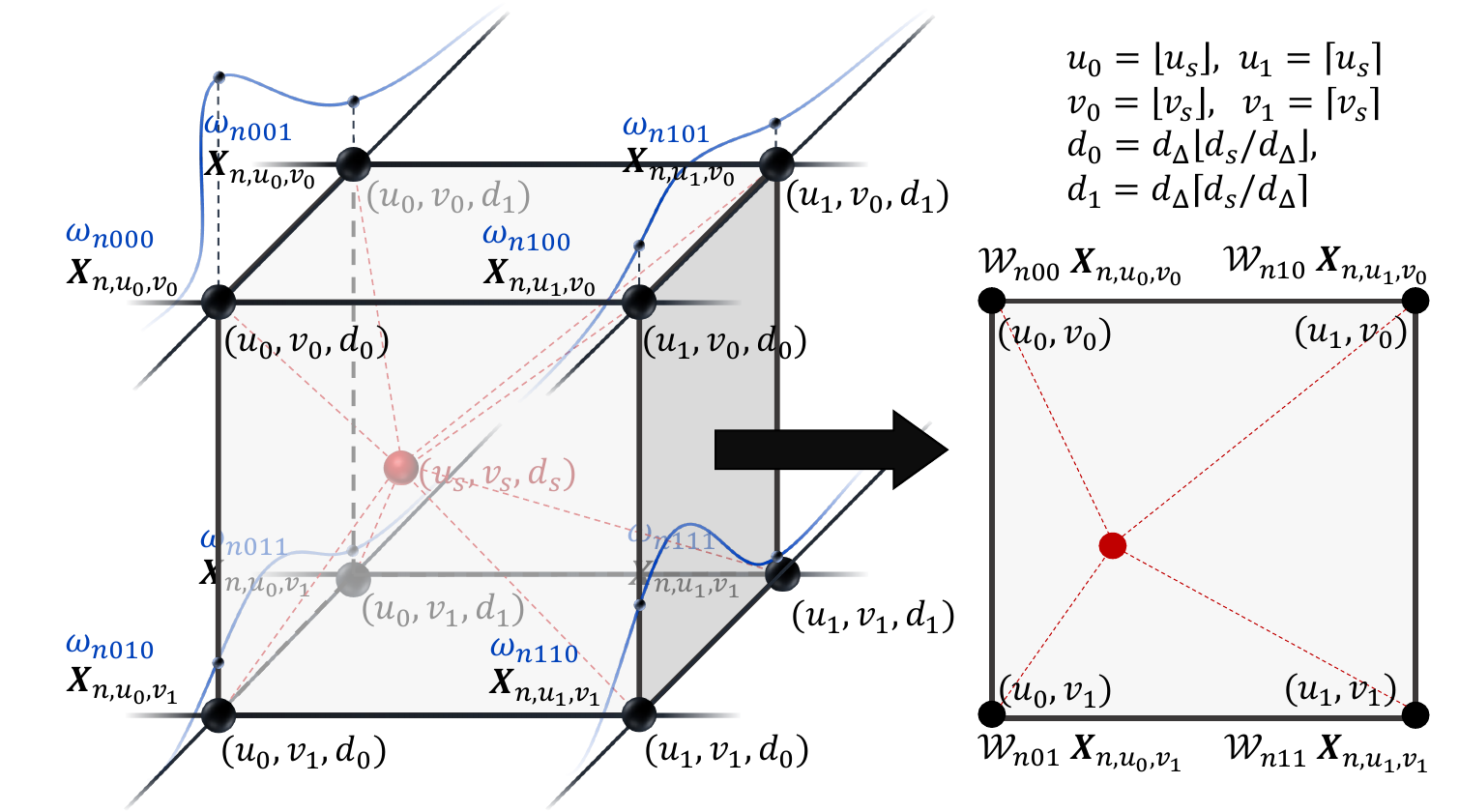}
    \caption{The original trilinear interpolation shown on the left can be simplified to a weighted bilinear interpolation as shown right.}
    \label{fig.conceptual_lift}
    \vspace{-3mm}
\end{figure}

As shown in Fig.~\ref{fig.conceptual_lift}, without loss of generality, for a specific view (the $n$-th view for example) with an expanded 3D image feature map $\boldsymbol{F}_n = \boldsymbol{D}_n \otimes_{-1} \boldsymbol{X}_n$, $\boldsymbol{F}_n \in \mathbb{R}^{H\times W\times D\times C}$, assume a sampling point $\boldsymbol{s}=(u_s,v_s,d_s)$ in the 3D pixel coordinate system falls in a cuboid formed by eight points: $\{(u_i, v_j, d_k)\}, i,j,k\in\left\{0,1\right\}$, which are the sampling point's eight nearest points in $\boldsymbol{F}_n$. The sampling point performs trilinear interpolation on the features of these eight points to obtain the sampling result. Since the eight points with the same $u,v$ coordinates share the same 2D image feature, their corresponding contributions in the interpolation process can be presented as 
$\left\{\omega_{nijk} \boldsymbol{X}_{n,u_i,v_j}\right\}$, where $\omega_{nijk}$ denotes $\boldsymbol{D}_{n,u_i,v_j,d_k}$ for notation simplicity. $\omega_{nijk}$ indicates the depth confidence scores of point $(u_i, v_j, d_k)$ selected from the discrete depth distributions $\boldsymbol{D}_{n,u_i,v_j}$ that it belongs to, according to its depth values $d_k$.
Therefore, the trilinear interpolation can be formulated as:
\begin{equation}
    \begin{aligned}
        &t_u = \frac{u_s - \lfloor u_s\rfloor}{\lceil u_s\rceil-\lfloor u_s\rfloor}, t_v = \frac{v_s - \lfloor v_s\rfloor}{\lceil v_s \rceil-\lfloor v_s\rfloor}, \\
        &t_d = \frac{d_s - \lfloor d_s/ d_{\Delta}\rfloor d_{\Delta}}{\lceil d_s/d_{\Delta}\rceil d_{\Delta} - \lfloor d_s/d_{\Delta}\rfloor d_{\Delta}}, \\
        &Trili(\boldsymbol{F}_n, 
\boldsymbol{s}) = \\
            &\omega_{n000} \left(1-t_d\right)  \left(1-t_u\right) \left(1-t_v\right) \boldsymbol{X}_{n,u_0,v_0} + \\
            &\omega_{n001} t_d      \left(1-t_u\right) \left(1-t_v\right) \boldsymbol{X}_{n,u_0,v_0} + \\
            &\omega_{n100} \left(1-t_d\right)  t_u     \left(1-t_v\right) \boldsymbol{X}_{n,u_1,v_0} + \\
            &\omega_{n101} t_d      t_u     \left(1-t_v\right) \boldsymbol{X}_{n,u_1,v_0} + \\
            &\omega_{n010} \left(1-t_d\right) \left(1-t_u\right)  t_v     \boldsymbol{X}_{n,u_0,v_1} + \\
            &\omega_{n001} t_d     \left(1-t_u\right)  t_v     \boldsymbol{X}_{n,u_0,v_1} + \\
            &\omega_{n110} \left(1-t_d\right)  t_u     t_v     \boldsymbol{X}_{n,u_1,v_1} + \\
            &\omega_{n111} t_d      t_u     t_v     \boldsymbol{X}_{n,u_1,v_1} ,
    \end{aligned}
    \label{equ.trilinear}
\end{equation}
where $d_\Delta$ indicates the depth interval between two adjacent depth bins. Eq. \ref{equ.trilinear} can be further re-formulated into
\begin{equation}
\begin{aligned}
    &\mathcal{W}_{nij}=\omega_{nij0}(1-t_d) + \omega_{nij1}t_d, \\
    &Trili(\boldsymbol{F}_n, 
    \boldsymbol{s}) = \\
            &\mathcal{W}_{n00} \left(1-t_u\right) \left(1-t_v\right) \boldsymbol{X}_{n,u_0,v_0} + \\
            &\mathcal{W}_{n10} t_u     \left(1-t_v\right) \boldsymbol{X}_{n,u_1,v_0} + \\
            &\mathcal{W}_{n01} \left(1-t_u\right) t_v     \boldsymbol{X}_{n,u_0,v_1} + \\
            &\mathcal{W}_{n11} t_u t_v \boldsymbol{X}_{n,u_1,v_1}.
\end{aligned}
\label{equ.trilinear2}
\end{equation}

Comparing Eq.~\ref{equ.trilinear} and Eq.~\ref{equ.trilinear2}, we can find that the trilinear interpolation here can be actually split into two parts: 1) a simple linear interpolation of the estimated discrete depth distribution along the depth axis to obtain the depth scores 
$\mathcal{W}_n=\left\{\mathcal{W}_{nij}\right\}$
, and 2) a depth score-weighted bilinear interpolation in the 2D image feature maps. Compared with bilinear interpolation, the only extra computation we need is to sample depth scores $\mathcal{W}_n$ through linear interpolation. With such an optimization, we can calculate the 3D deformable attention on the fly rather than maintaining all the expanded features. Thus the entire attention mechanism can be implemented in a more efficient way. Moreover, such an optimization can reduce half of the multiplication operations and thus speedup the overall computation as well. 

We conduct a simple experiment to compare the efficiency. The results are shown in Table~\ref{tab:source_consume}. For a fair comparison, we show the resource consumption of all steps in detail. 
The results show that our efficient implementation only takes about 3\% time cost and 1\% memory consumption compared with the vanilla one.  

\begin{table}[h]
\centering
\caption{The comparisons of resource consumption (Memory / Time) between the vanilla one and our efficient one. ``Expand'', ``Aggregate'' and ``Full'' indicate feature expanding, feature aggregation, and overall process respectively. Our efficient one does not need ``Expand''.}

\resizebox{0.47\textwidth}{!}{
\renewcommand\arraystretch{1.2}
\begin{tabular}{l|c|c|c} 
\shline
Method                             & Expand            & Aggregate  & Full \\
\shline  
Vanilla            & 25303MB / 76.5ms                    & 674MB / 85.4ms            & 3204MB / 161.9ms\\
\shline
Efficient         & - / -                              & 29MB / 5.2ms             & 29MB / 5.2ms  \\
\shline
\end{tabular}
}
\vspace{2pt}
\label{tab:source_consume}
\end{table}

\subsection{Analysis}
\label{subsec.analysis}
\noindent\textbf{Comparison with Lift-Splat-based Feature Lifting.} Lift-Splat-based feature lifting follows the pipeline of LiDAR-based methods. To construct pseudo LiDAR features as if they are obtained from LiDAR data, it also utilizes depth estimation to supplement depth information. 
It obtains the pseudo LiDAR features by constructing single-scale 3D expanded feature maps explicitly and transforming them from a 3D pixel coordinate system into the ego coordinate system according to the camera parameters. The pseudo LiDAR features are further assigned to their nearest 3D anchors to generate lifted features for downstream tasks. Since the locations of both 3D anchors and pseudo LiDAR features are constant, the assignment rule between them is fixed based on their geometrical relationship. 

In DFA3D, the relationship between 3D queries and the expanded 3D features that are computed based on the estimated sampling locations can also be considered as an assignment rule. Instead of being fixed, DFA3D can progressively refine the assignment rule by updating the sampling locations layer by layer in a Transformer-like architecture.
Besides, the efficient implementation of DFA3D enables the utilization of multi-scale 3D expanded features, which are more crucial to detecting small (faraway) objects.

\noindent\textbf{Comparison with 2D Attention-based Feature Lifting.} Similar to DFA3D, 2D  attention-based feature lifting also transforms each 3D query into a pixel coordinate system. However, to satisfy the input of off-the-shelf 2D attention operators, a 3D query needs to be projected to a 2D one, where its depth information is discarded. It is a compromise to implementation and can cause multiple 3D queries to collapse into one 2D query. Hence multiple 3D queries can be entangled greatly. This can be proved easily according to the pin-hole camera model. In such a scenario, according to Eq.~\ref{equ.sameuv}, as long as a 3D query's coordinate $(x,y,z)$ in the ego coordinate system satisfies 

\begin{equation}
\begin{aligned}
    \frac{x}{z} = \frac{u_n-u_n^0}{f_n^x}, \frac{y}{z} = \frac{v_n-v_n^0}{f_n^y}
\end{aligned},
\label{equ.view_ray}
\end{equation}
its reference point in the target 2D pixel coordinate system will be projected to the shared $(u_n, v_n)$ coordinate, resulting in poorly differentiated lifted features that will be used to update the 3D queries' content and downstream tasks.

\section{Experiments}
We conduct extensive experiments to evaluate the effectiveness of DFA3D-based feature lifting compared with other feature lifting methods. We integrate DFA3D into several open-source methods to verify its generalization ability and portability.  

\subsection{Experiment Setup}

\paragraph{Dataset and Metrics}
We follow previous works~\cite{li2022bevformer, li2022bevdepth, wang2022detr3d} to conduct experiments on the nuScenes dataset~\cite{caesar2020nuscenes}. The nuScenes dataset contains 1,000 sequences, which are captured by a variety of sensors (e.g. cameras, RADAR, LiDAR). Each sequence is about 20 seconds, 
and is annotated in 2 frames/second. Since we focus on the camera-based 3D detection task, we use only the image data taken from 6 cameras. 
There are 40k samples with 1.4M annotated 3D bounding boxes in total. We use the classical splits of training, validation, and testing set with 28k, 6k, and 6k samples, respectively.
We follow previous works~\cite{li2022bevformer, li2022bevdepth, wang2022detr3d} to take the official evaluation metrics, including mean average precision (mAP), ATE, ASE, AOE, AVE, and AAE for our evaluations. 
These six metrics evaluate results from center distance, translation, scale, orientation, velocity, and attribute, respectively. 
In addition, a comprehensive metric called nuScenes detection score (NDS) is also provided.

\vspace{-0.5cm}
\paragraph{Implementation Details}
For fair comparisons, when implementing DFA3D on the open-source methods, we keep all configurations the same without bells and whistles.
For all experiments, without specification, we take BEVFormer-base~\cite{li2022bevformer} as our baseline, which uses a ResNet101-DCN~\cite{he2016deep,dai2017deformable} backbone initialized from FCOS3D~\cite{wang2021fcos3d} checkpoint and a Transformer with 6 encoder layers and 6 decoder layers. 
All experiments are trained with 24 epochs using AdamW optimizer~\cite{adamw} with a base learning rate of $2\times 10^{-4}$ on 8 NVIDIA Tesla A100 GPUs.

\subsection{Comparisons of Feature Lifting Methods}
To compare different feature lifting methods fairly, we set feature lifting as the only variable and keep the others the same. More specifically, we block the temporal information and utilize the same 2D backbone and 3D detection head. The results are shown in Table~\ref{tab:operator}.

We first compare four different feature lifting methods in the first four rows in Table~\ref{tab:operator}. All models use one layer for a fair comparison. The results show that {\methodname} outperforms all previous works, which demonstrates the effectiveness. 
Although the larger receptive field makes 2D deformable attention (DFA2D)-based method obtains much better result than the point attention-based method, its performance is not satisfactory enough compared with the Lift-Splat-based one. One key difference between DFA2D and Lift-Splat is their depth utilization. The extra depth information enables the Lift-Splat-based method to achieve a better performance, although the depth information is predicted and not accurate enough. 
Different from the DFA2D-based feature lifting, DFA3D enables the usage of depth information in the deformable attention mechanism and helps our DFA3D-based feature lifting achieve the best performance.

As discussed in Sec.~\ref{subsec.analysis}, the adjustable assignment rule enables attention-based methods to conduct multi-layer refinement. With the help of one more layer refinement, DFA2D-based feature lifting surpasses Lift-Splat. However, benefitting from the better depth utilization, our DFA3D-based feature lifting still maintains the superiority.

\begin{table}[t]
\centering
\renewcommand\arraystretch{1.2}
\caption{Comparisons of feature lifting methods. DFA2D denotes the 2D deformable attention. 
We use the same architecture and supervision for depth estimation as our {\methodname} for the Lift-Splat-based method. We also leverage SECOND FPN~\cite{yan2018second} to enable multi-scale feature maps for the Lift-Splat-based method.
}
\resizebox{1.01\linewidth}{!}{
\setlength{\tabcolsep}{1.5pt}
\begin{tabular}{l|c|c|ccccc|c} 
\shline
\textbf{Method}            & \textbf{\#layers} &
\textbf{mAP}$\uparrow$  & \textbf{mATE}$\downarrow$ & \textbf{mASE}$\downarrow$  & \textbf{mAOE}$\downarrow$ & \textbf{mAVE}$\downarrow$ & \textbf{mAAE}$\downarrow$ & \textbf{NDS}$\uparrow$ \\
\shline
PointAttn & 1            & 35.0 & 76.3 & 27.8  & 42.9  & \textbf{84.9} & \textbf{20.7} & 42.2  \\
DFA2D & 1          & 35.9 & 74.6 & 27.8  & 42.5  & 85.8 & 20.9 & 42.8  \\
Lift-Splat      & 1                    & 36.9 & \textbf{73.1} & 28.0  & 44.7  & 85.0 & 23.6 & 43.0  \\
 \rowcolor{gray!15} {\methodname} (Ours) & 1          & \textbf{37.3} & 73.4 & \textbf{27.5}  & \textbf{41.7}  & 85.2 & 22.5 & \textbf{43.6}  \\
\shline
DFA2D & 2          & 37.1 & 72.7 & \textbf{27.7}  & 43.5  & 81.4 & \textbf{20.6} & 44.0  \\
 \rowcolor{gray!15} {\methodname} (Ours) & 2     & \textbf{37.9} & \textbf{72.4} & \textbf{27.7}  & \textbf{38.1}  & \textbf{78.1} & 22.3 & \textbf{45.1}  \\
\shline
\end{tabular}
}
\vspace{2pt}

\label{tab:operator}
\end{table}

\begin{table}[t]
\centering
\renewcommand\arraystretch{1.2}
\caption{Comparisons of our method and baselines. Without specifications, all models use ResNet101~\cite{he2016deep} as the backbone. 
$^\dag$ denotes that the model adopts ResNet101-DCN~\cite{dai2017deformable} as the backbone. 
Sparse4D$^*$ denotes the Sparse4D without Depth Reweight Module (DRM) proposed by~\cite{lin2022sparse4d}.
}
\resizebox{\linewidth}{!}{
\setlength{\tabcolsep}{1.5pt}
\begin{tabular}{l|c|ccccc|c} 
\shline
\textbf{Method}              & 
\textbf{mAP}$\uparrow$  & \textbf{mATE}$\downarrow$ & \textbf{mASE}$\downarrow$  & \textbf{mAOE}$\downarrow$ & \textbf{mAVE}$\downarrow$ & \textbf{mAAE}$\downarrow$ & \textbf{NDS}$\uparrow$ \\
\shline
DETR3D$^\dag$~\cite{wang2022detr3d}           & 34.7 & 76.5 & 26.8 & \textbf{39.2}   & 87.6 & 21.1 & 42.2  \\
\rowcolor{gray!15} DETR3D-\methodname$^\dag$            & \textbf{35.5(+0.8)} & \textbf{74.4} & \textbf{26.8} & 41.6   & \textbf{86.4} & \textbf{20.7} & \textbf{42.8(+0.6)}  \\
\shline
BEVFormer-t~\cite{li2022bevformer}               & 25.2 & 90.0 & 29.4 & 65.5   & 65.7 & 21.6 & 35.4  \\
\rowcolor{gray!15} BEVFormer-t-\methodname& \textbf{26.9(+1.7)} & \textbf{88.0} & 29.2 & \textbf{60.6}   & \textbf{60.6} & 23.1 & \textbf{37.3(+1.9)}  \\
\shline
BEVFormer-s~\cite{li2022bevformer}                  & 37.0 & 72.1 & 28.0 & \textbf{40.7}   & 43.6 & 22.0 & 47.9  \\
\rowcolor{gray!15} BEVFormer-s-\methodname      & \textbf{40.1(+3.1)} & \textbf{72.1} & \textbf{27.9} & 41.1   & \textbf{39.1} & \textbf{19,6} & \textbf{50.1(+2.2)}  \\
\shline
BEVFormer-b$^\dag$~\cite{li2022bevformer}             & 41.6 & 67.3 & 27.4 & 37.2   & 39.4 & \textbf{19.8} & 51.7  \\
\rowcolor{gray!15} BEVFormer-b-\methodname  $^\dag$    & \textbf{43.0(+1.4)} & \textbf{65.4} & \textbf{27.1} & 37.4   & \textbf{34.1} & 20.5 & \textbf{53.1(+1.4)}  \\
\shline
DA-BEV-S$^\dag$~\cite{dabev}        & 42.8 & \textbf{63.3} & 27.3 & 33.1   & 32.7 & \textbf{18.8} & 53.9  \\
\rowcolor{gray!15} DA-BEV-S-\methodname $^\dag$       & \textbf{43.5(+0.7)} & 64.5 & \textbf{27.0} & \textbf{32.8}   & \textbf{31.6} & 20.2 & \textbf{54.2(+0.3)}  \\
\shline
DA-BEV$^\dag$~\cite{dabev}        & 43.3 & \textbf{62.3} & 27.1 & 35.1   & 30.9 & \textbf{18.8} & 54.5  \\
\rowcolor{gray!15} DA-BEV-\methodname $^\dag$       & \textbf{44.1(+0.8)} & 62.6 & \textbf{27.4} & \textbf{33.4}   & \textbf{31.1} & 19.4 & \textbf{54.7(+0.2)}  \\
\shline
Sparse4D$^*$$^\dag$~\cite{dabev}        & 43.2 & - & - & 37.9   & - & - & 53.7  \\
\rowcolor{gray!9} Sparse4D$^*$-DRM$^\dag$~\cite{dabev}    & 43.6(+0.4) & 63.3 & 27.9 & \textbf{36.3}   & 31.7 & \textbf{17.7} & 54.1(+0.4)  \\
\rowcolor{gray!15} Sparse4D$^*$-\methodname $^\dag$       & \textbf{44.6(+1.4)} & \textbf{63.2} & \textbf{27.1} & 38.8   & \textbf{30.5} & 18.1 & \textbf{54.5(+0.8)}  \\
\shline
\end{tabular}
}

\label{tab:generalization_method_size}
\vspace{-8pt}
\end{table}

\subsection{Generalizations for Different Models}
To verify the generalization and portability of DFA3D-based feature lifting, we evaluate it on various open-sourced methods that rely on 2D deformable attention or point attention-based feature lifting. 
Thanks to the mathematical simplification of DFA3D, our DFA3D-based feature lifting can be easily integrated into these methods with only a few code modifications (please refer to the appendix for a more intuitive comparison).

We compare baselines to those that integrate DFA3D-based feature lifting in Table~\ref{tab:generalization_method_size}. 
The results show that DFA3D brings consistent improvements in different methods, indicating its generalization ability across different models. Furthermore, DFA3D introduces significant gains of $1.7$, $3.1$, and $1.4$ mAP on BEVFormer-t\footnote{We use the BEVFormer-t, BEVFormer-s, and BEVFormer-b for the BEVFormer-tiny, BEVFormer-small, and BEVFormer-base variants in \url{https://github.com/fundamentalvision/BEVFormer}.}, BEVFormer-s, and BEVFormer-b, respectively, showing the effectiveness of DFA3D across different model sizes. Note that, for a fair comparison, we use one sampling point and fixed sampling offsets when conducting experiments on DETR3D~\cite{wang2022detr3d}.

For a more comprehensive comparison, we also conduct experiments on two concurrent works DA-BEV~\cite{dabev} and Sparse4D~\cite{lin2022sparse4d}, who design their own modules to address the depth ambiguity problem implicitly or through post-refinement. 
Differently, we solve the problem from the root at the feature lifting process, which is a more principled solution. The results show that, even with their efforts as a foundation, DFA3D achieves $+0.7$ mAP, $+0.8$ mAP and $+1.0$ mAP improvement over DA-BEV-S, DA-BEV~\cite{dabev} and Sparse4D~\cite{lin2022sparse4d} respectively, which verifies the necessity of our new feature lifting approach.

\subsection{Ablations}
\paragraph{Mixture of 3D and 2D Deformable Attention-based Feature Lifting.}
We verify the effectiveness of our DFA3D-based feature lifting by mixing DFA2D-based and DFA3D-based feature lifting in the encoder of BEVFormer. 
As shown in Table~\ref{tab:num_layer}, in the first $N_{2D}$ layers we use the DFA2D-based feature lifting, and in the following $N_{3D}$ layers we use the DFA3D-based one.
The results show a consistent improvement with the usage of DFA3D-based feature lifting increases. The progressive improvements indicate that the more DFA3D introduced, the less depth ambiguity problem is, which finally results in a better performance.

\begin{table}[t]
\centering
\renewcommand\arraystretch{1.2}
\caption{The effect of different numbers of DFA3D-based feature lifting used in the BEVFormer's encoder layers. $N_{2D}$ and $N_{3D}$ are the numbers of layers that use DFA2D-based and DFA3D-based feature lifting respectively.}
\vspace{1.5pt}
\resizebox{0.9\linewidth}{!}{
\setlength{\tabcolsep}{1.5pt}
\begin{tabular}{c|c|c|ccccc|c} 
\shline
\textbf{$N_{2D}$}            & \textbf{$N_{3D}$}   &
\textbf{mAP}$\uparrow$  & \textbf{mATE}$\downarrow$ & \textbf{mASE}$\downarrow$  & \textbf{mAOE}$\downarrow$ & \textbf{mAVE}$\downarrow$ & \textbf{mAAE}$\downarrow$ & \textbf{NDS}$\uparrow$ \\
\shline
4 & 2    & 42.0 & 66.9 & 27.1 & 35.3  & 34.2 & 19.1 & 52.7  \\
2 & 4    & 42.7 & 66.0 & 27.5 & 38.0  & 32.6 & 20.2 & 52.9 \\
0 & 6    & 43.0 & 65.4 & 27.1 & 37.4  & 34.1 & 20.5 & 53.1
\\
\shline
\end{tabular}
}
\label{tab:num_layer}
\end{table}

\begin{table}[t]
\centering
\renewcommand\arraystretch{1.3}
\caption{The effects of DepthNet module and depth quality. ``Depth Sup.'' means that only supervise the DepthNet but do not use the estimated depth in feature lifting. ``Unsup.'', ``Sup.'', and ``GT'' denote the depth used in feature lifting, that is unsupervised learned, supervised learned, and ground truth (from LiDAR). }
\vspace{1.5pt}
\resizebox{\linewidth}{!}{
\setlength{\tabcolsep}{1.5pt}
\begin{tabular}{c|l|c|ccccc|c} 
\shline
\textbf{Row}            & 
\textbf{Method}            & 
\textbf{mAP}$\uparrow$  & \textbf{mATE}$\downarrow$ & \textbf{mASE}$\downarrow$  & \textbf{mAOE}$\downarrow$ & \textbf{mAVE}$\downarrow$ & \textbf{mAAE}$\downarrow$ & \textbf{NDS}$\uparrow$ \\
\shline
1 & BEVFormer                     & 41.6 & 67.3 & 27.4 & 37.2 & 39.4 & 19.8 & 51.7  \\
2 & BEVFormer-Depth Sup.           & 42.0 & 66.9 & 27.4 & 37.6 & 38.6 & 19.9 & 52.1  \\
\shline
3 & BEVFormer-\methodname\ Unsup.  & 41.7 & 67.3 & 28.0 & 36.7   & 35.3 & 19.6 & 51.8  \\
4 & BEVFormer-\methodname\ Sup.    & 43.0 & 65.4 & 27.1 & 37.4   & 34.1 & 20.5 & 53.1  \\
\shline
5 & BEVFormer-\methodname\ GT     & 56.7 & 45.3 & 26.8 & 24.7   & 32.8 & 19.2 & 62.4  \\
\shline
\end{tabular}
}
\label{tab:depth_sup_and_quality}
\end{table}

\paragraph{Effects of DepthNet Module and Depth Quality.}
As shown in Table~\ref{tab:depth_sup_and_quality}, we first present the influence of the DepthNet module.
Compared with the 1.4\% mAP gain (Row 4 vs. Row 1) brought by DFA3D, simply equipping BEVFormer with DepthNet and supervising it brings in a 0.4\% mAP gain (Row 2 vs. Row 1), which is relatively marginal.
The comparison indicates that the main improvement does not come from the additional parameters and supervision of DepthNet, but from a better feature lifting.

We then evaluate the influence of depth qualities. We experiment with three different depth, 1) learned from unsupervised learning, 2) learned from supervised learning, and 3) ground truths generated from LiDAR. The corresponding results of these different depth qualities are available in Rows 3, 4, and 5 in Table~\ref{tab:depth_sup_and_quality}. The results show that the performance improves with better depth quality. Remarkably, the model variant with ground truth depth achieves 56.7\% mAP, outperforming baselines by 15.1\% mAP. It indicates a big improvement space which deserves a further in-depth study of obtaining better depth estimation.

\paragraph{Effects of Feature Querying Methods.}
Deformable attention enables querying features dynamically. It generates sampling points according to the queried features from the previous layer. Inspired by previous work~\cite{wang2022detr3d}, we freeze the position of sampling points, under which the attention operation actually degenerates to the trilinear interpolation. We compare the degenerated one with the default one in Table~\ref{tab:onepointordeform}. The results show the effectiveness of the deformable operation. The dynamic sampling can help focus on more important positions and enable better feature querying.
\vspace{-0.15cm}

\begin{table}[t]
\centering
\renewcommand\arraystretch{1.2}
\caption{Comparison of different feature querying methods. Trilinear indicates querying features on the expanded 3D feature maps only at the reference points. }
\vspace{1.5pt}
\resizebox{\linewidth}{!}{
\setlength{\tabcolsep}{1.5pt}
\begin{tabular}{l|c|ccccc|c} 
\shline
\textbf{Transformation}            & 
\textbf{mAP}$\uparrow$  & \textbf{mATE}$\downarrow$ & \textbf{mASE}$\downarrow$  & \textbf{mAOE}$\downarrow$ & \textbf{mAVE}$\downarrow$ & \textbf{mAAE}$\downarrow$ & \textbf{NDS}$\uparrow$ \\
\shline
Trilinear       & 40.6 & 68.7 & 27.7  & 38.9  & 35.5 & 18.9 & 51.3  \\
DFA3D   & 43.0 & 65.4 & 27.1  & 37.4  & 34.1 & 20.5 & 53.1  \\
\shline
\end{tabular}
}
\label{tab:onepointordeform}
\vspace{-1.5mm}
\end{table}

\begin{table}[t]
\centering
\renewcommand\arraystretch{1.3}
\caption{The evaluation of efficiency and resource consumption of {\methodname} based on BEVFormer. }
\vspace{2pt}
\resizebox{0.75\linewidth}{!}{
\setlength{\tabcolsep}{1.5pt}
\begin{tabular}{l|c|c|c} 
\shline
\textbf{Method}            & 
\textbf{Speed (FPS)} & \textbf{GPU Mem (GB)} & \textbf{\#Param (M)}\\
\shline
BEVFormer                     & 2.8 & 6.9 & 62.1 \\
\shline
BEVFormer-\methodname               & 2.4 & 7.0 & 68.6 \\
\shline
\end{tabular}
}
\vspace{2pt}
\label{tab:efficiency}
\vspace{-5mm}
\end{table}
\paragraph{Efficiency and resource consumption.}
We evaluate the efficiency and resource consumption of DFA3D based on BEVFormer, as shown in Table~\ref{tab:efficiency}. DFA3D-based feature lifting requires a little more resources and is slightly slower. We clarify that the side effects, especially the extra parameters mainly come from DepthNet, which is adjustable and can be optimized with the development of depth estimation.

\section{Conclusion}

In this paper, we have presented a basic operator called 3D deformable attention (DFA3D), built upon which we develop a novel feature lifting approach. Such an approach not only takes depth into consideration to tackle the problem of depth ambiguity but also benefits from the multi-layer refinement mechanism. 
We seek assistance from math to simplify DFA3D and develop a memory-efficient implementation. The simplified DFA3D makes querying features in 3D space through deformable attention possible and efficient. The experimental results show a consistent improvement, demonstrating the superiority and generalization ability of DFA3D-based feature lifting.

\textbf{Limitations and Future Work.} In this paper, we simply take monocular depth estimation to provide depth information for DFA3D, which is not accurate and stable enough. Recently proposed methods~\cite{sparse4dv2, streampetr} have proved that long temporal information can provide the network with a better depth sensing ability. How to make full use of the superior depth sensing ability to generate high-quality depth maps explicitly, and utilize them to help improve the performance of DFA3D-based feature lifting is still an issue, which we leave as future work.
{\small
\bibliographystyle{ieee_fullname}
\bibliography{egbib}
}

\clearpage

\onecolumn
\section*{Appendix}

\appendix

\section{Closer visualization of the expanded 3D image features.}
\begin{figure}[h]
    \centering
	\includegraphics[width=1\linewidth]{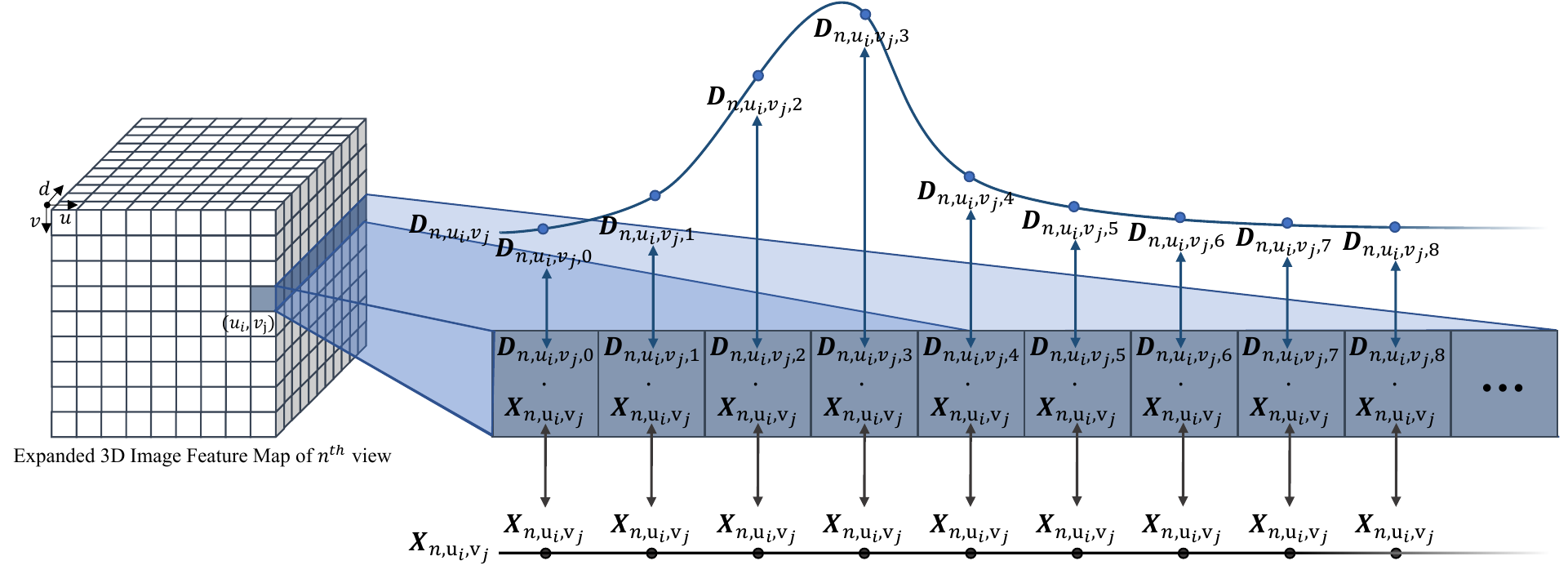}
    \caption{A closer visualization of the expanded 3D image features.}
    \label{fig.further_explanation}
    \vspace{-3mm}
\end{figure}

In Sec.~\ref{subsec.feature_expanding}, for a specific view (the $n^{th}$ view for example), we obtain the expanded 3D image feature map by conducting outer product at last dimension between the estimated depth distribution $\boldsymbol{D}_{n} \in \mathbb{R}^{H\times W\times D}$ and 2D image feature maps $\boldsymbol{X}_n \in \mathbb{R}^{H\times W\times C}$. To further explain the resulting expanded 3D image feature $\boldsymbol{F}_n \in \mathbb{R}^{H\times W\times D\times C}$, we show a closer visualization in Fig.~\ref{fig.further_explanation}. For the features with the same $(u_i, v_j)$ coordinate (zoomed in), they share the same 2D image feature  $\boldsymbol{X}_{n,u_i,v_j}$ and depth distribution $\boldsymbol{D}_{n,u_i,v_j}$. However, since they have different depth values ${d_k}$, they select different depth confidence scores ${\boldsymbol{D}_{n,u_i,v_j,d_k}}$, resulting in different features $\{\boldsymbol{D}_{n,u_i,v_j,d_k}\cdot\boldsymbol{X}_{n,u_i,v_j}\}$.

\section{Can 3D deformable attention be trivially implemented by feature weighting followed with a common 2D deformable attention?}
\begin{figure}[h]
    \centering
	\includegraphics[width=1\linewidth]{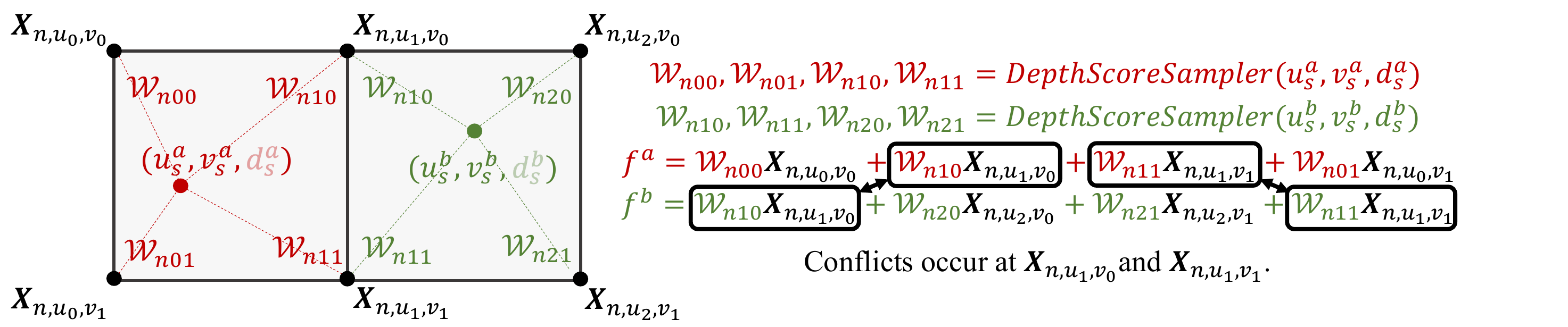}
    \caption{In 3D deformable attention, when sampling features for two sampling points $(u_s^a, v_s^a, d_s^a)$ and $(u_s^b, v_s^b, d_s^b)$ respectively, we first sample depth scores $(\mathcal{W}_{n00}, \mathcal{W}_{n10}, \mathcal{W}_{n11}, \mathcal{W}_{n01})$ and $(\mathcal{W}_{n10}, \mathcal{W}_{n20}, \mathcal{W}_{n21}, \mathcal{W}_{n11})$ for these two points respectively. After that, we conduct depth-weighted bilinear interpolation for these two points based on their own depth scores independently.}
    \label{fig.on_the_fly}
    \vspace{-3mm}
\end{figure}

As each sampling point in 3D deformable attention has its own 3D location, it will lead to its own depth score for 4 adjacent image features when conducting weighted bilinear interpolation. 
Inevitably, there will be features ($\boldsymbol{X}_{n,u_1,v_0}$ and $\boldsymbol{X}_{n,u_1,v_1}$ in Fig.~\ref{fig.on_the_fly}) referred by more than one sampling point when sampling points are located in neighboring grids.  In such cases, feature weighting will results in conflicts. Thus, we can not simply prepare 2D features by depth-weighting and then conduct the common 2D deformable attention. The depth-weighted feature computation should be conducted on the fly. 

\newpage
\section{Applicability}
\begin{figure}[h]
    \centering
	\includegraphics[width=0.95\linewidth]{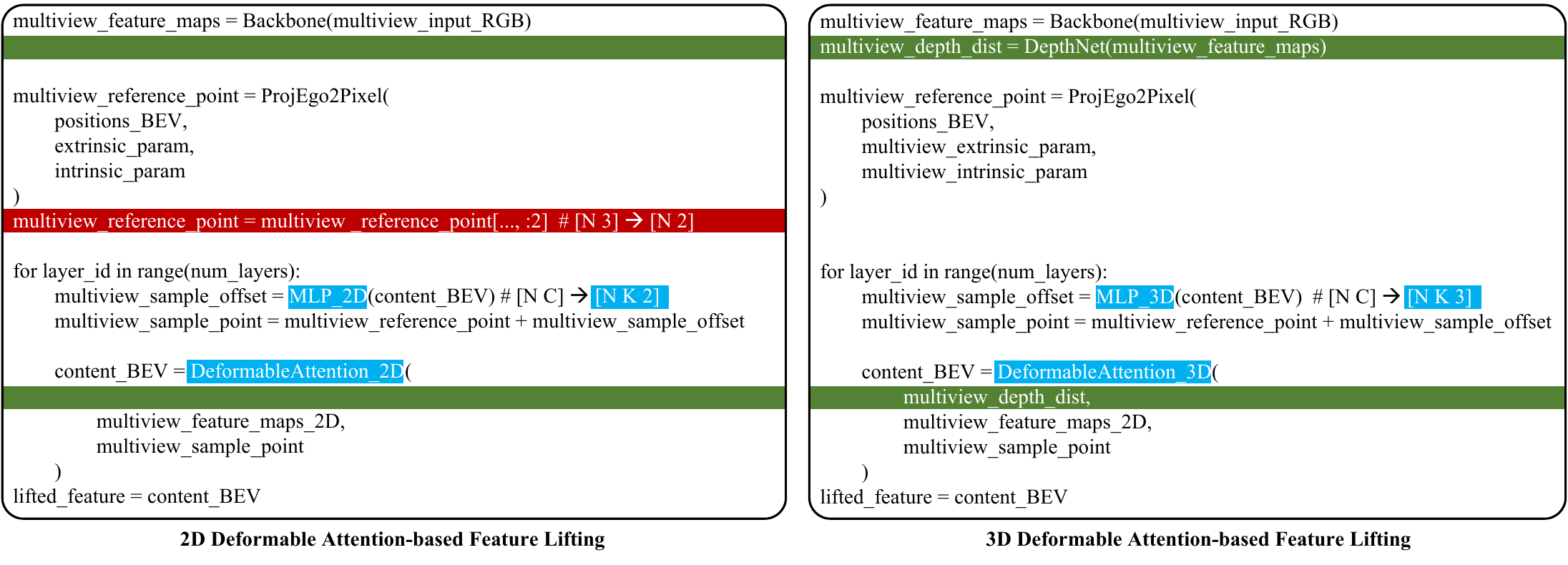}
    \caption{Integrate {\methodname} into any 2D deformable attention-based feature lifting requires only a few modifications in code.}
    \label{fig.code_compare}
    \vspace{-3mm}
\end{figure}

As the comparison shown in Fig.~\ref{fig.code_compare}, when integrating our {\methodname} in any 2D deformable attention-based feature lifting only requires a few modifications in code. The main modification lies in the addition of DepthNet and replacing 2D deformable attention with our 3D deformable attention.

\section{Visualization}
\begin{figure*}[h]
    \centering
    \includegraphics[width=0.95\linewidth]{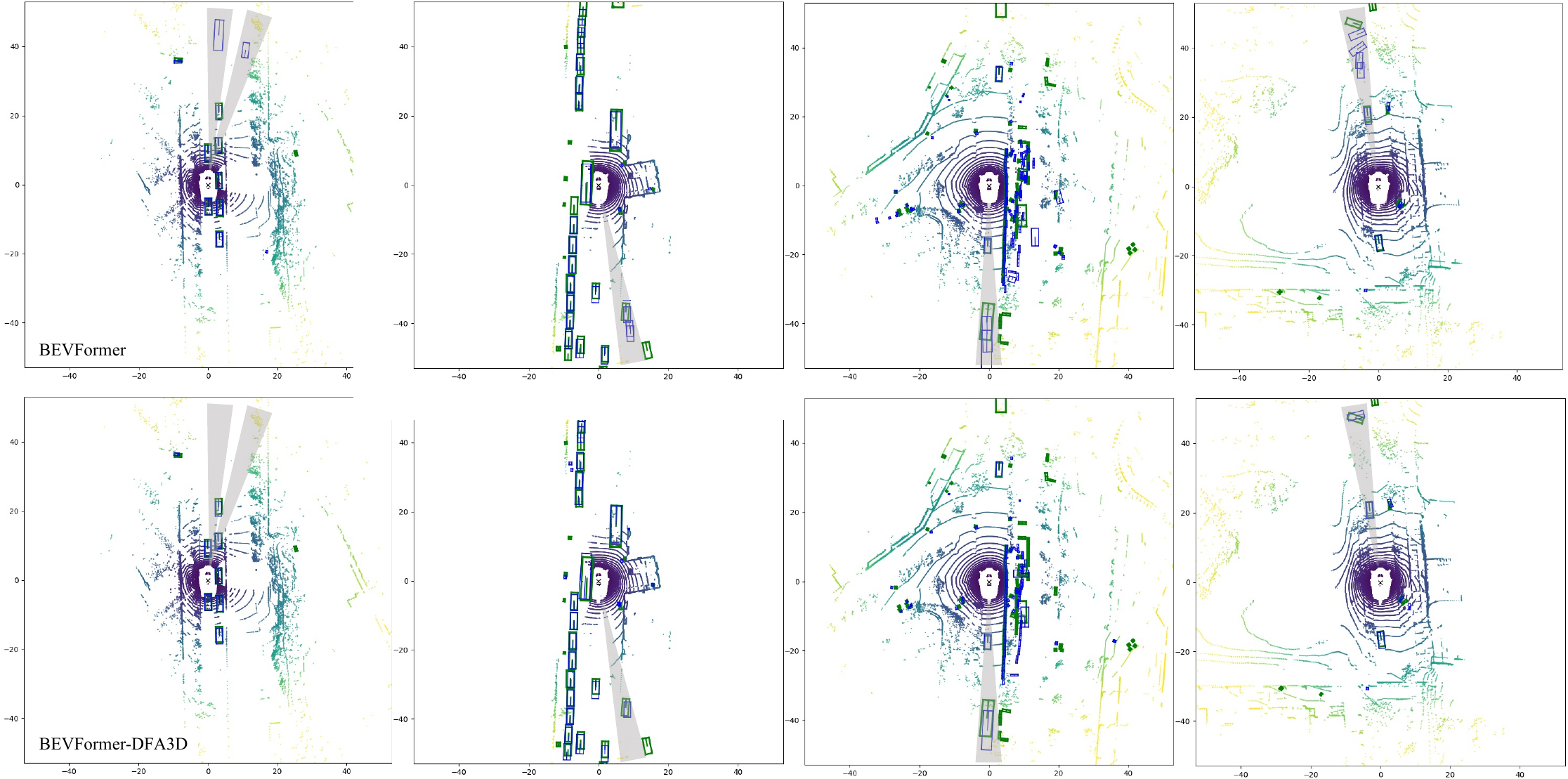}
    \caption{The visualization of predictions in BEV using green boxes to represent ground truth boxes and blue boxes to represent predicted ones. Duplicate predictions, which are caused by the depth ambiguity problem, are enclosed by shaded triangles.}
    \label{fig.visualization}
    \vspace{-1mm}
\end{figure*}

We visualize the predictions of BEVFormer and BEVFormer-{\methodname} in Fig.~\ref{fig.visualization}. The shaded triangles correspond to camera rays, in which BEVFormer makes more duplicate predictions behind or in front of ground truth objects compared with BEVFormer-{\methodname}.
It demonstrates the negative effects of depth ambiguity in 2D deformable attention. 
After integrating with \methodname, the wrong predictions caused by the depth ambiguity problem are reduced.

\end{document}